\documentclass[10pt,twocolumn,letter]{article}
\usepackage{titling}
\usepackage{ijcb}
\usepackage{times}
\usepackage{epsfig}
\usepackage{graphicx}
\usepackage{amsmath}
\usepackage{amssymb}
\usepackage{gensymb}
\usepackage{tabularx}
\usepackage{algorithm}
\usepackage{algpseudocode}
\usepackage{comment}
\usepackage{caption}
\usepackage{subcaption}

\usepackage{xcolor}

\newcommand{\W}{\mathcal{W}}
\newcommand{\Z}{\mathcal{Z}}
\newcommand{\dir}[1]{\mathbf{\hat{n}_{#1}}}

\usepackage[pagebackref=false,breaklinks=true,letterpaper=true,colorlinks,bookmarks=false]{hyperref}

\ijcbfinalcopy 

\ifijcbfinal\fi
\begin{document}
\title{\vspace{-0.5cm}On the use of automatically generated synthetic image datasets for benchmarking face recognition\vspace{-0.5cm}}
\author{Laurent Colbois
\and Tiago de Freitas Pereira
\and Sébastien Marcel\\
Idiap Research Institute, Martigny, Switzerland\\
{\tt\small \{laurent.colbois, tiago.pereira, sebastien.marcel\}@idiap.ch}
}
\maketitle

\thispagestyle{empty}

\begin{abstract}
The availability of large-scale face datasets has been key in the
progress of face recognition. 
However, due to licensing issues or copyright infringement, some datasets are
not available anymore (e.g. MS-Celeb-1M).
Recent advances in Generative Adversarial Networks (GANs), to synthesize
realistic face images, provide a pathway to replace real datasets by synthetic datasets,
both to train and benchmark face recognition (FR) systems. 
The work presented in this paper provides a study on benchmarking FR systems using
a synthetic dataset. 
First, we introduce the proposed methodology to generate a synthetic dataset,
without the need for human intervention, by exploiting the latent structure of
a StyleGAN2 model with multiple controlled factors of variation.
Then, we confirm that (i) the generated synthetic identities are not data subjects from the GAN's training
dataset, which is verified on a synthetic dataset with 10K+ identities;
(ii) benchmarking results on the synthetic dataset are a good substitution, often providing error
rates and system ranking similar to the benchmarking on the real dataset.
\vspace{-0.5cm}
\end{abstract}



\section{Introduction}
\begin{figure}
	\centering
	\includegraphics[width=1.0\linewidth]{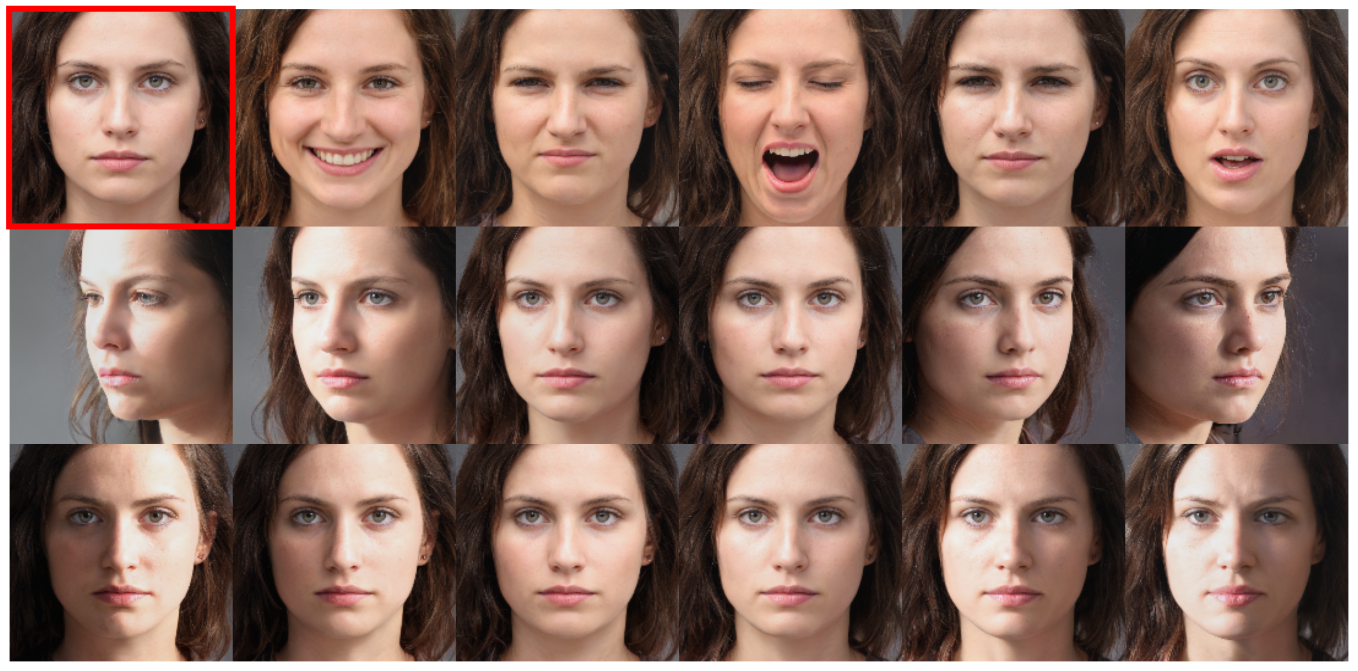}
	\caption{Example of the full set of variations generated for a single
	synthetic identity. \textbf{All displayed images are synthetic.} The
	first image (highlighted in red) is the main reference (neutral
	expression, frontal view, frontal illumination). Then, using a latent
	editing approach we \textbf{automatically} generate expression
	variations (first row), pose variations (second row) and illumination
	variations (third row).}
	\label{fig:variationsexample}
	\vspace{-0.5cm}
\end{figure}

Face datasets used to train and benchmark face recognition (FR) systems have
reached the scale of hundreds of thousand identities. These large-scale
datasets have been responsible for pushing the state-of-the-art face
recognition systems in terms of recognition rates. Many of them claim to be based on
images of ``celebrities'' scraped from the web. However, this is partially inaccurate as these
datasets are not only composed of public figures from media, but may include as well
ordinary people~\cite{murgia_whos_2019} uninformed about the use of their data. 
As a consequence, well established datasets had to be discontinued~\cite{harvey_exposingai_nodate}. 
In parallel, legislation is trying to address these issues. For instance, in Europe the GDPR~\cite{gdpr} has
established that biometric data is personal data and requires that informed
consent is obtained from data subjects.
Hence, it becomes harder to collect, curate and distribute face recognition
datasets that comply with all regulations such as anonymity or informed
consent.
At the same time, the past few years have seen major improvements in synthetic
face image generation, in particular using Generative Adversarial Networks (GANs). 
The most recent GANs enable easy sampling of high-resolution and very realistic
face images. Moreover, their latent space present some structure that can be
exploited to edit face attributes in a disentangled manner, and thus to generate
many different samples of the same synthetic identity. 

The aim of this work is to explore if those synthetic images can be
substituted to real data when benchmarking FR systems. In a past study on this
topic~\cite{sumi_study_2005}, three required characteristic of such an
evaluation dataset have been identified: 
(i)~\textbf{Privacy}: each biometric data in the synthetic dataset should not
represent any real person,
(ii)~\textbf{Precision}: the evaluation results derived from a synthetic
biometric dataset should be equal to the one from the real dataset and
(iii)~\textbf{Universality}: the precision requirement should be satisfied for
all evaluated authentication algorithms. 
In the present work, we will mainly assess whether the \emph{precision} and
\emph{privacy} requirements are satisfied with our synthetic dataset. The
\emph{universality} requirement is much more difficult to evaluate. We cover it partially in this work by considering several systems, but it will be more extensively addressed in future work.

\begin{figure}
    \centering
    \includegraphics[width=0.7\linewidth]{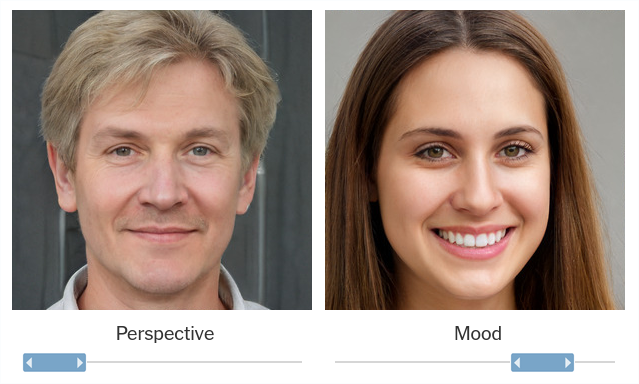}
    \vspace{-0.3cm}
    \caption{Usual methods to find editing directions do not provide insight on the scale of the editing. Human intervention is required at generation time to explore how much one can move along a direction while preserving identity and realism of the image, which is unpractical for automatic dataset generation. (Illustration from \cite{hill_designed_2020})}
    \label{fig:sliders}
    \vspace{-0.4cm}
\end{figure}

This paper first describes our dataset generation method, which relies on
exploiting semantic editing directions in the latent space of a face-based
StyleGAN2~\cite{stylegan2}. Then we will check whether the synthetic
identities are actually new or if they are similar to real identities from the
GAN's training dataset. We will also compare the performance of several FR
systems on precise protocols using real and synthetic datasets independently. 
Finally, we will inspect some of the synthetic samples causing
recognition errors to identify possible shortcomings of our method. 
Overall, our main contributions are the following:
\vspace{-0.1cm}
\begin{itemize}
	\item We introduce a new method for finding semantic editing
	directions in the latent space of StyleGAN2, based on the projection
	in this space of a real dataset with labeled covariates.
	This method enables \textbf{fully automatic} editing, unlike current semantic
	editing methods (cf. fig. \ref{fig:sliders}) 
    \vspace{-0.2cm}
	\item We create a synthetic dataset imitating the Multi-PIE dataset \cite{multipie},
	called \emph{Syn-Multi-PIE}, to investigate if it can
	be substituted to a real dataset when evaluating the robustness of FR systems
	to illumination, expression and pose variations. We
	hypothesize that this assessment can give an insight about the
	adequacy of synthetic datasets to benchmark FR systems.
\end{itemize}

\section{Related work}
\subsection{Synthetic face generation}
Since the early introduction of generative adversarial networks (GANs) in \cite{goodfellow_generative_2014}, their generative ability has often been showcased on the task of face generation. The current state-of-the-art model is the StyleGAN2 architecture trained on the FFHQ dataset  \cite{stylegan2}, which provides the most perceptually compelling face images, while also performing best regarding the Fréchet Inception Distance (FID), the most common metric used to quantify realism of the generation. However, the StyleGAN2 model is unconditioned : while it allows to sample face images from random input latent vectors, it does not provide \emph{a priori} control on the semantic content of the result (e.g. pose, expression or race of the person). It is notably non trivial to generate several variations of the same identity. Several approaches work towards overcoming this limitation.

First, one can use image-to-image translation networks to generate variations through \emph{a posteriori} editing, the same way one would edit a real image. We can mention in particular \cite{he_attgan_2019} and \cite{liu_stgan_2019}, which propose methods for editing a broad set of facial attributes, but there exists many references proposing an editing method targeting a specific attribute such as age \cite{antipov_face_2017} or expression \cite{ding_exprgan_2018}.

A second approach is instead to retrain the generative model to make it conditional. This is generally done by ensuring the latent input to the generator can be split between a component specifying identity and a component specifying all other factors of variations (\cite{donahue_semantically_2017}, \cite{trigueros_generating_2018}, \cite{bao_towards_2018}), with sometimes a even finer control on the factors of variation to manipulate several semantic attributes in a disentangled manner (\cite{marriott_ivi_gan}, \cite{GIF2020}). 

Finally, it is also possible to exploit StyleGAN2's properties. In \cite{nitzan_face_2020}, the authors propose to decouple disentanglement from synthesis by learning a mapping into StyleGAN2's latent space, from which they are able to generate face images by respectively mixing the identity and attributes of two input synthetic images, while keeping the synthesis quality. In \cite{shen_interpreting_2020}, the authors exploit the property of linear separation of semantic attributes in StyleGAN2's latent space, and propose a general framework to find latent directions corresponding to editing only a particular semantic attribute, while keeping other attributes untouched. In a sense, this makes the unconditional model actually conditional, by leveraging latent properties of the model that were learned in an unsupervised manner.

\subsection{Synthetic data in face recognition}

Usage of synthetic data in face recognition usually occurs at the training step, as a form of data augmentation.  A distinction can be made between depth augmentation (increasing the number of samples per subject) and width augmentation (increasing the number of subjects). Depth augmentation is typically done by mapping face images to a 3D morphable model (3DMM), enabling generation of pose, expression or illumination augmentations \cite{crispell_dataset_2017}. For width augmentation, it is also possible to use similar 3DMMs by randomizing their parameters \cite{kortylewski_training_2018}. GAN based approaches can be exploited both for depth- and width- augmentation, although with them identity preservation in depth-augmentation is a major challenge. We can mention \cite{trigueros_generating_2018}, in which they train an identity-disentangled GAN and use it to augment a FR train dataset, and \cite{marriott_3d_2020}, where they design a 3D GAN conditioned on the parameters of a 3DMM to improve facial recognition on extreme poses.

In this work however, we focus on the use of synthetic data to benchmark face recognition systems. We are not aware of any recent work on this topic. We can mention \cite{sumi_study_2005} that proposes a procedure to replace a real evaluation dataset with a synthetic one that produces similar biometric score distributions, but their work completely predates recent progresses in GAN-based face generation.

\section{Synthetic generation methodology}
For the face generator, we use the StyleGAN2 model pretrained on FFHQ \cite{stylegan2}. The network architecture decomposes in two parts. First, a \emph{mapping} network, which is a fully-connected network, takes Gaussian-sampled latent vectors as input and transforms them into a new latent of same dimension. Its input and output spaces are respectively called $\Z$ and $\W$. Secondly, a \emph{synthesis} network uses $\W$-latents as input style vectors used to control the actual face generation process. The output of this network is a face image. One can easily generate a database with many independent faces by randomly sampling latent vectors. However, in order to use this synthetic data in face recognition applications, we need to generate several images where the identity is preserved but other attributes can vary. In this work, we focus on three types of variability~: illumination, pose and expression. This is done by exploiting the remarkable property  of \emph{linear separability} of StyleGAN2's $\W$ space.  This property emerges as a byproduct of the training of the GAN. Its impact is that when considering binary semantic attributes characterizing a face (young/old, female/male, neutral/smiling expression...), we can usually find a hyperplane in $\W$ that separates well the $\W$-latents from each class. The normal to this hyperplane can then be used as a latent direction that specifically edits the considered covariate, thus providing finer control on the generation, despite the network being unconditional in the first place. In this section, we describe our protocol for constructing a face recognition dataset by exploiting the linear separability of $\W$.

\subsection{Finding latent directions}
We take inspiration from \cite{shen_interpreting_2020}, which proposes an approach to find semantic latent directions by labeling two distinct populations of $\W$-latents vectors according to a binary target attribute (e.g left vs right profile), then fitting a linear SVM in $\W$ to separate those two populations. In their case, the latent vectors are randomly sampled then labeled by running their associated image through an auxiliary pretrained attribute classification network. Their method has the partial drawback of needing this auxiliary network, but more importantly it does not provide any notion of \textbf{scaling} of the editing, i.e. how much along one latent direction one can move while preserving the identity and the realism of the image.
To avoid those issues, we propose an alternative approach for obtaining the latent populations on which the SVMs are fitted. Instead of using random images, we \emph{project} into the $W$ space the content of a dataset of real images. By projection is meant the process of finding a $\W$-latent for which the associated image is perceptually close to a target real image. 
This is done through optimization, generally using the perceptual loss as the objective to minimize. Several variants of this approach exists (\cite{rolux_roluxstylegan2encoder_2021}, \cite{abdal_image2stylegan_2019}), mainly differing in the details of the optimization process. We use the implementation from \cite{stylegan2}.

 This projection-based approach removes the need for an auxiliary classification network, and it gives access to a sense of \textbf{scale} : as a reasonable heuristic, we can keep track of the \emph{average distance to the hyperplane} of each population of projected latents. We hypothesize that by using this distance as the editing scale for our synthetic identities, the resulting range of variation for each attribute will be similar to the one observed in the real dataset. This enables us to generate all variations of each synthetic identity in a \textbf{fully automatic} manner, without requiring human intervention to decide the strength of each editing. 

\begin{figure}
	\centering
	\includegraphics[width=0.9\linewidth]{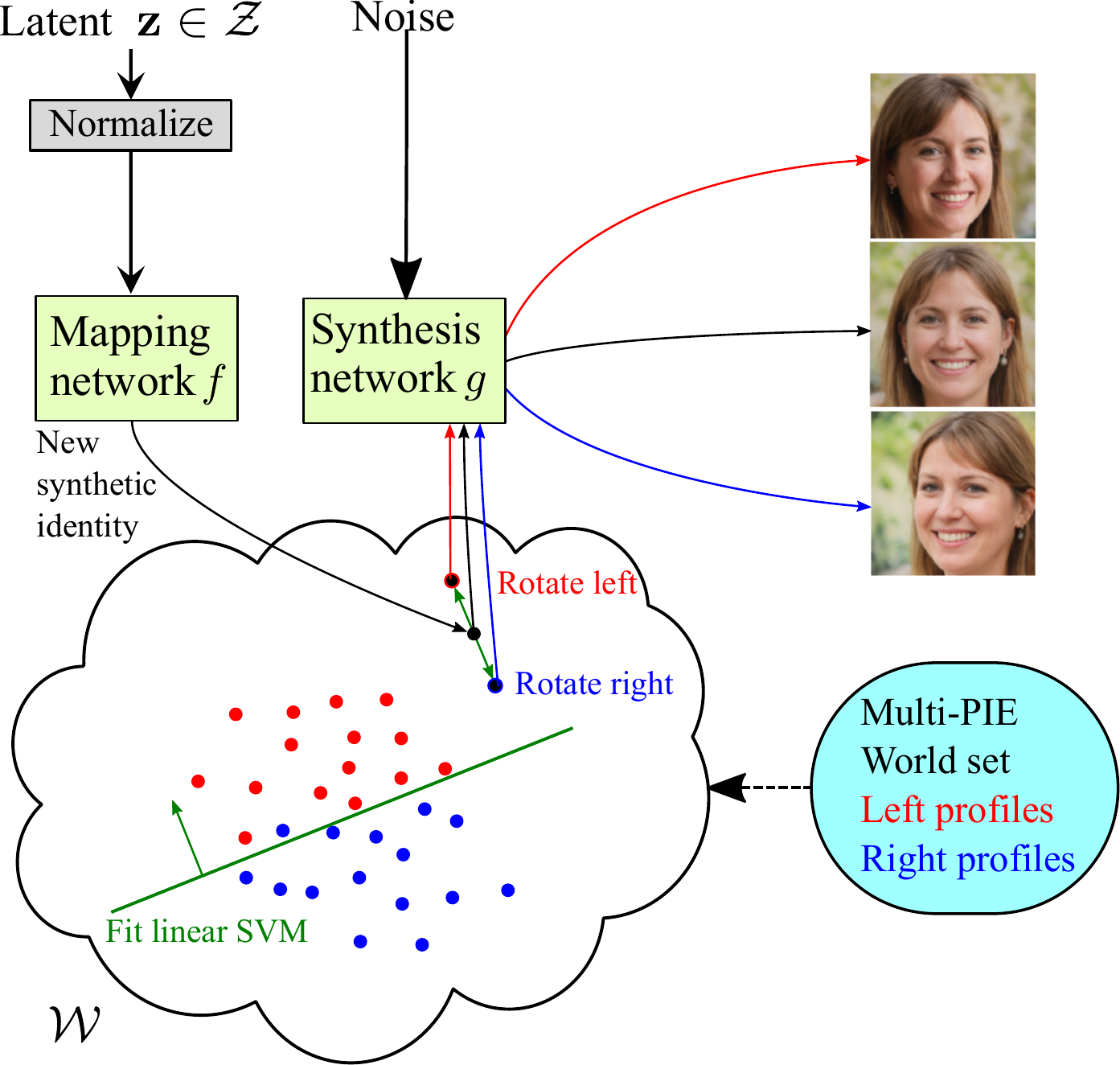}
	\caption{To obtain a latent direction, we project in $\W$ labeled images, then fit a linear SVM to separate the latents. The normal to the resulting hyperplane is used as the editing direction for this attribute. New identities are obtained by randomly sampling a $\Z$-latent. We then edit its associated $\W$-latent using our computed latent directions to obtain variations of the original image. 
	}
	\label{fig:latentdirectionsschema}
	\vspace{-0.5cm}
\end{figure}

We project the Multi-PIE dataset \cite{multipie}. For each identity, it contains face images with labeled covariates for pose, illumination and expression. The available expressions are \emph{neutral, smile, disgust, scream, squint, surprise}. After projection, we use the resulting latents to fit SVMs in $W$ and find the following interesting latent directions: left-right pose edition, left-right illumination edition, and edition between any pair of expressions, along with the associated maximum editing scales. Figure \ref{fig:latentdirectionsschema} illustrate the full process.

\subsection{Syn-Multi-PIE Dataset}

Finally, we generate a dataset of synthetic faces with controlled variations. New identities are obtained by sampling a latent in the $\Z$ space, and we create the main reference image by neutralizing the face in the $\W$ space, which means we edit it to ensure it has frontal view, frontal lightning and neutral expression. For pose and illumination neutralization, we simply project the $\W$-latent onto the separation hyperplane. For expression neutralization, a preliminary qualitative observation of the typical output of StyleGAN2 shows that randomly sampled latents mostly produce neutral or smiling expressions. We thus simply edit the latent along the neutral-to-smile direction to increase its neutral component. 

To ensure enough variability between identities, we optionally apply a constraint on the interclass threshold (ICT), which specifies the minimum face embedding distance required between any identity pair. We only accept new candidate identities if they satisfy the constraint w.r.t every previous identity. The embedding distance is measured using a Inception-ResNet-v2 model \cite{inceptionresnet} pretrained on Ms-Celeb \cite{msceleb}. In a second pass, we generate variations for each identity using our precomputed latent directions. For scaling our edits,  we use the measured average distance of Multi-PIE projections to the hyperplanes as the maximal variation. Figure \ref{fig:variationsexample}
presents example of the full set of variations obtained for a given face. We will showcase in the following sections that this synthetic database satisfies the \emph{privacy} and \emph{precision} requirements, making it a good substitution to Multi-PIE to benchmark FR systems. We thus name it \textbf{Syn-Multi-PIE}. With a concern for reproducibility of our work, we also release the code to regenerate this database.\footnote{\url{https://gitlab.idiap.ch/bob/bob.paper.ijcb2021_synthetic_dataset}} Additional details on the generation algorithm can also be found in the supplementary material.

\section{Are StyleGAN2 identities new ?}\label{sec:uniqueness}
To assess the requirement of \emph{privacy}, we need to verify that generated identities do not simply reproduce existing identities from the FFHQ dataset. We evaluate this by reproducing on our synthetic dataset an experiment originally proposed in \cite{varkarakis_validating_2020} on the first version of StyleGAN. It consists in comparing the identity similarity between a synthetic dataset (Sy) and a seed dataset (Se), which is the dataset used to train the face generator. A third dataset (Ref) of real data  with labeled identities is used as reference to estimate the density of lookalikes in a standard population.

\begin{figure}
	\centering
	\includegraphics[width=0.8\linewidth]{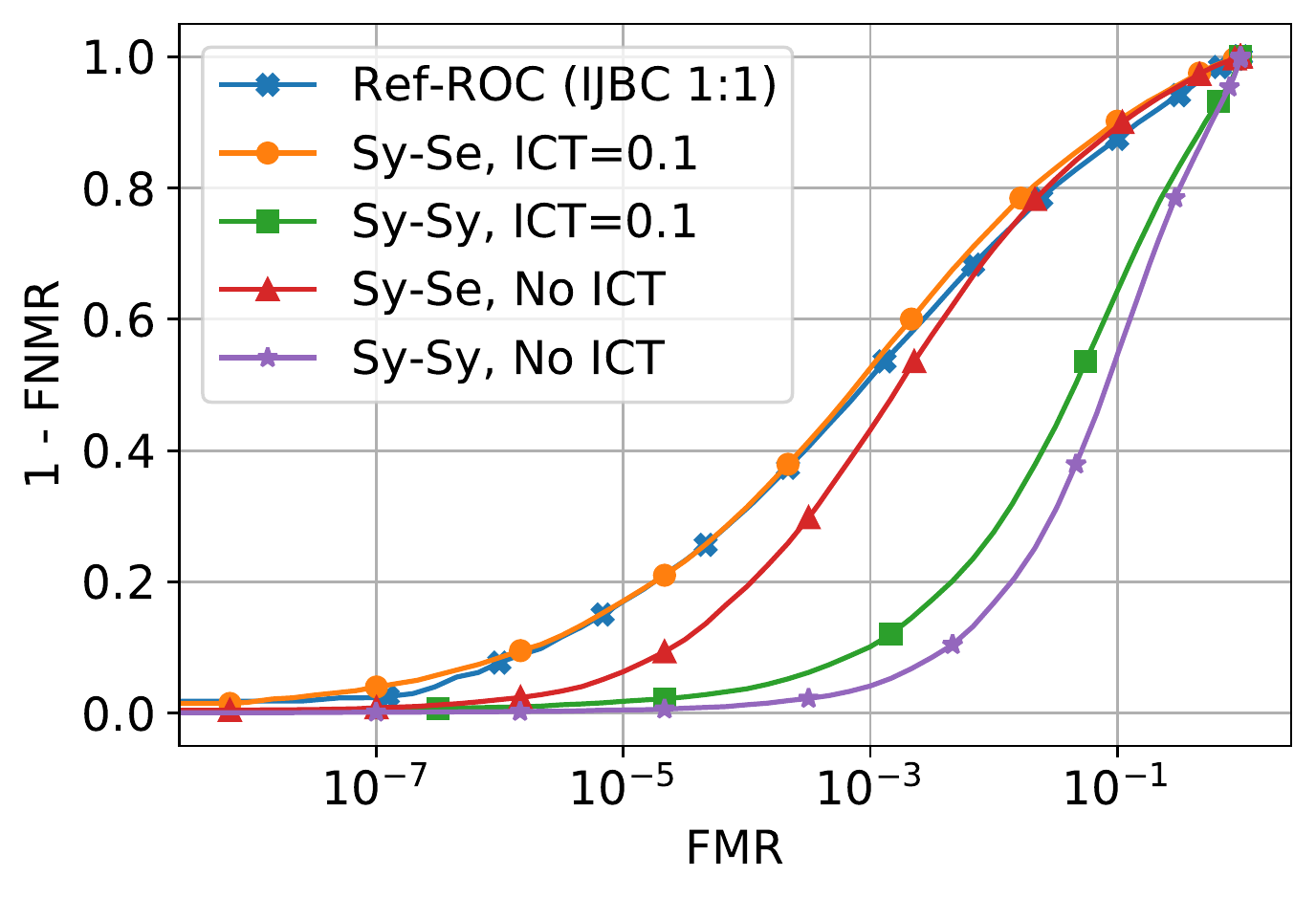}
	\vspace{-0.4cm}
	\caption{ROC curves obtained when comparing identities from the synthetic dataset (Sy) to the seed dataset (FFHQ). The Ref-ROC is obtained using the IJB-C dataset (with the 1:1 verification protocol).}
	\label{fig:uniqueness}
	\vspace{-0.2cm}
\end{figure}

For this experiment we use an Inception-ResNet v2 trained on Casia-WebFace \cite{casiawebface} for face embedding extraction. We measure face similarity as the negative cosine distance between face embeddings. 
We then compare three ROC curves. The Ref-ROC is built solely from identities from the reference dataset. For the Sy-Se-ROC, we keep genuine pairs from Ref, and assuming that Sy identities should all be different from Se ones, we use Sy-Se comparisons as impostor pairs. If this ROC curve is shifted downwards with respect to the Ref-ROC, it means that Sy-Se pairs contain more false matches than Ref impostors pairs, i.e. Sy-Se pairs contains more lookalikes than the Ref dataset, and thus Sy identities are not truly new. If however the Sy-Se ROC is superposed to the Ref-ROC, then we can say Sy identities are new. Doing the same experiment with Sy-Sy pairs, we can also assess the amount of lookalikes \emph{between} Sy identities. For establishing the Ref-ROC, we work with the IJB-C dataset under the 1:1 verification protocol \cite{ijbc}. We generate two sets of 11k synthetic identities, once with no ICT constraint, and once with an ICT constraint of 0.1. The value of 0.1 was empirically chosen as a good-trade off between identity variability, and number of rejected faces during the generation.
Those 11k identities are compared with the 70k images of the seed dataset (FFHQ) to obtain the Sy-Se curves, and they are compared between themselves to get the Sy-Sy curves. The resulting ROC curves are presented in figure \ref{fig:uniqueness}.

\begin{figure}
	\centering
	\includegraphics[width=0.8\linewidth]{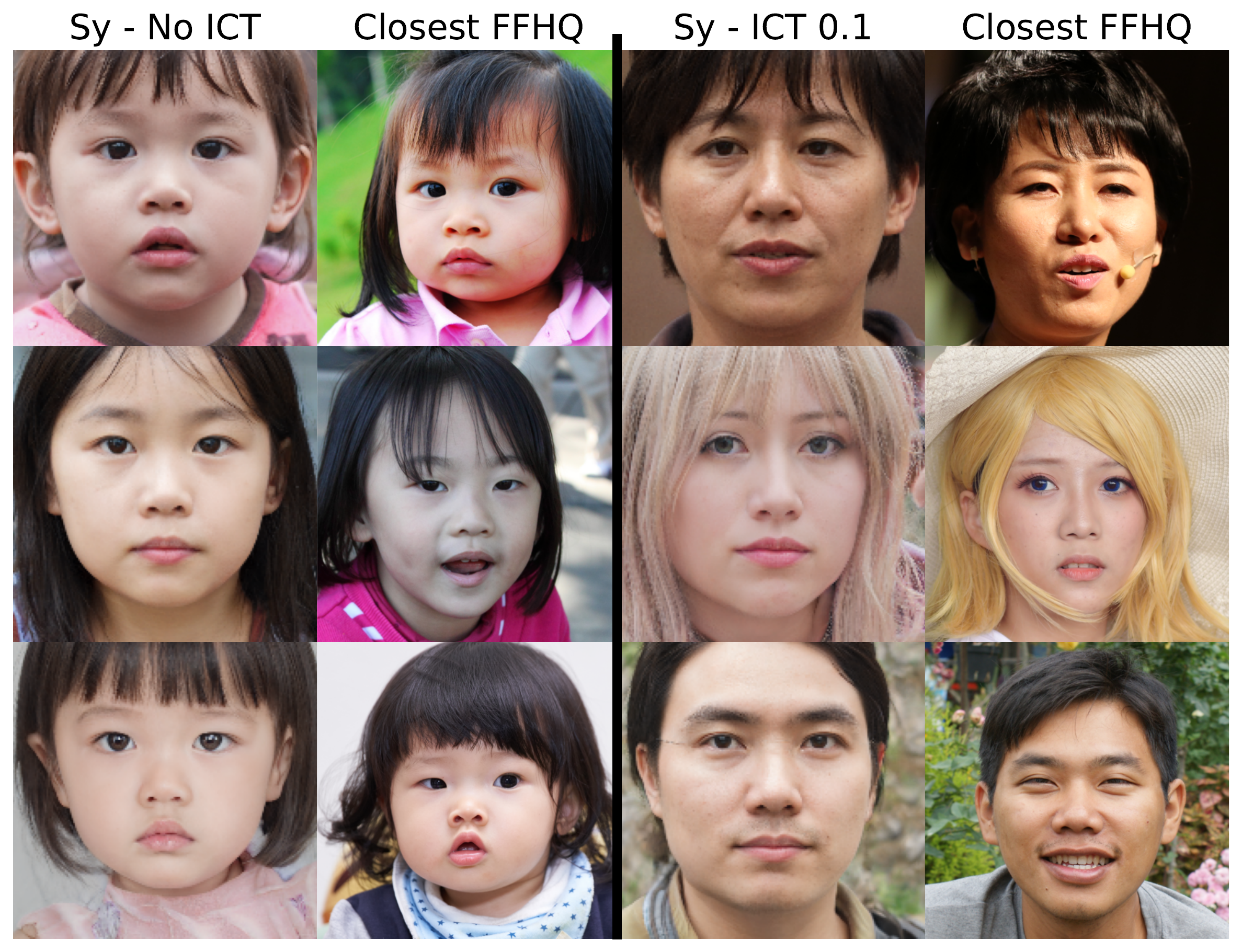}
	\vspace{-0.2cm}
	\caption{Closest matches in FFHQ to synthetic identities, with no ICT constraint (leftmost columns) and with an ICT constraint of 0.1 (rightmost columns).}
	\label{fig:ffhqclosestmatches}
    \vspace{-0.6cm}
\end{figure}

We first focus on the Sy-Se curves. We observe that without an ICT constraint, the Sy-Se-ROC lies below the Ref-ROC. This indicates there is a higher density of lookalikes between Sy and Se than in a real population, which is here modeled by IJB-C. This is similar to the observation made in \cite{varkarakis_validating_2020} using StyleGAN. In the article, they notice that this is caused by the presence of children in FFHQ, and thus also in StyleGAN's output distribution, while SOTA face recognition networks are typically not trained on faces of children and thus perform poorly on this population. We observe the same behavior, as showcased in the leftmost columns of figure \ref{fig:ffhqclosestmatches} that present the closest Sy-Se matches. They solved the issue by manually removing the children, but this is unpractical if we plan to generate a large number of identities. Alternatively, we can apply the ICT constraint. Indeed, this greatly reduces the amount of children in the synthetic dataset: many children candidate faces are rejected by the model used to apply the ICT constraint, as this model itself fails at distinguishing different children. We showcase on the rightmost columns of figure \ref{fig:ffhqclosestmatches} how introducing the ICT constraint impacts the closest Sy-Se lookalikes: while children are mostly removed, the lookalikes now appear to be from an east-asian population. We can hypothesize that this FR model might still be underperforming this demographic, maybe due to it being underrepresented in the training data. However,
a study on the demographic differentials of this model is out of the scope of this work. Despite this, our Sy-Se-ROC is now quite well superposed to the Ref-ROC. This suggests there is the same density of lookalikes between the Sy and Se database as in IJB-C - at least under the scrutiny of this particular FR model. Therefore the Sy identities are as ``novel" as possible. Focusing on the Sy-Sy curves, we observe however that the variability of identities \emph{inside} the synthetic database is lesser than in real data, but this does not invalidate the results of the Sy-Se experiment that shows that the generator is indeed not just reproducing existing FFHQ identities. This is an important point which validates the use of synthetic face data as a way to satisfy the \emph{privacy} requirement.

\section{Can we use a synthetic dataset for benchmarking face recognition systems ?}
The central question of this work is whether we can completely replace real, sensitive FR data by synthetic and private one in a \emph{benchmark} setup. We thus have to assess the \emph{precision}  requirement, i.e. verify whether we can do this real-to-synthetic substitution and still obtain similar conclusions on the performance of several FR systems (error rate, system ranking and robustness to different factors of variation). We hypothesize that the answer to this question could depend on the range of considered variations, and on the quality of our face editing process, which might be better for some covariates than for others. For this reason, we choose not to perform a large-scale experiment which would not enable a fine analysis. Instead, we reproduce synthetically 3 Multi-PIE evaluation protocols, each of which targets a single covariate, and we compare the performance on the equivalent synthetic and real protocols. The U protocol targets illumination (enrollment with frontal lightning, probing with lightning variations), the E protocol targets expression (enrollment with neutral expression, probing with other expressions), and the P protocol targets pose (enrollment with frontal view, probing with other views). Equivalent protocols with Syn-Multi-PIE are also defined, using the generated variations shown in figure \ref{fig:variationsexample}. We always use the reference image for the enrollment, and probe the system only with variations of the target covariate. We use the same number of 64 identities in both setups. 6 FR systems, listed in the caption of table \ref{tab:fnmr}, are then benchmarked on those protocols.

\subsection{Experiment results}

\begin{table}[t]
	\begin{center}
		\begin{tabular}{|l|ccc|ccc|}
			\hline
			Database & \multicolumn{3}{c|}{Multi-PIE} & \multicolumn{3}{c|}{Syn-Multi-PIE} \\
			Protocol & U & E & P & U & E & P  \\
			\hline\hline
            Gabor                       &     0.42 &  0.46 &  0.66 &  0.43 &  0.46 &  0.67 \\
            LGBPHS                              &     0.43 &  0.42 &  0.72 &   \textbf{0.30} &  0.38 &  0.67 \\
            AF                 &     0.11 &   0.3 &  0.55 &  0.12 &  0.25 &  0.51 \\
            Inc-Res.v1 &     0.05 &  0.14 &   0.50 &  0.09 &   \textbf{0.50} &  0.47 \\
            Inc-Res.v2 &     0.07 &  0.15 &  0.26 &  0.08 &  \textbf{0.45} &  \textbf{0.44} \\
            AF-VGG2                    &     0.08 &  0.16 &   0.40 &   0.10 &  \textbf{0.44} &  \textbf{0.47} \\
			\hline
		\end{tabular}
	\end{center}
	\vspace{-0.5cm}
	\caption{FNMR values at FMR@1E-3. We highlighted all the scores where the error rate on Syn-Multi-PIE was more that 5\% away from the error rate on the equivalent Multi-PIE protocol. The considered systems are a Gabor graph (\emph{Gabor}) \cite{Gunther_frice}, a local Gabor binary pattern histogram sequence (\emph{LGBPHS}) \cite{Gunther_frice}, and 4 neural net based systems : original ArcFace (\emph{AF}) \cite{arcface}, ArcFace retrained on VGGFace2 (\emph{AF-VGG2}) \cite{cao_vggface2_2018}, Inception-ResNet models trained on Casia-WebFace (\emph{Inc-Res.v1} and \emph{v2}) \cite{pereira_heterogeneous_2019}.}
	\label{tab:fnmr}
	\vspace{-0.6cm}
\end{table}
\begin{figure}
	\centering
	\begin{subfigure}[b]{0.45\linewidth}
		\centering
		\includegraphics[width=\textwidth]{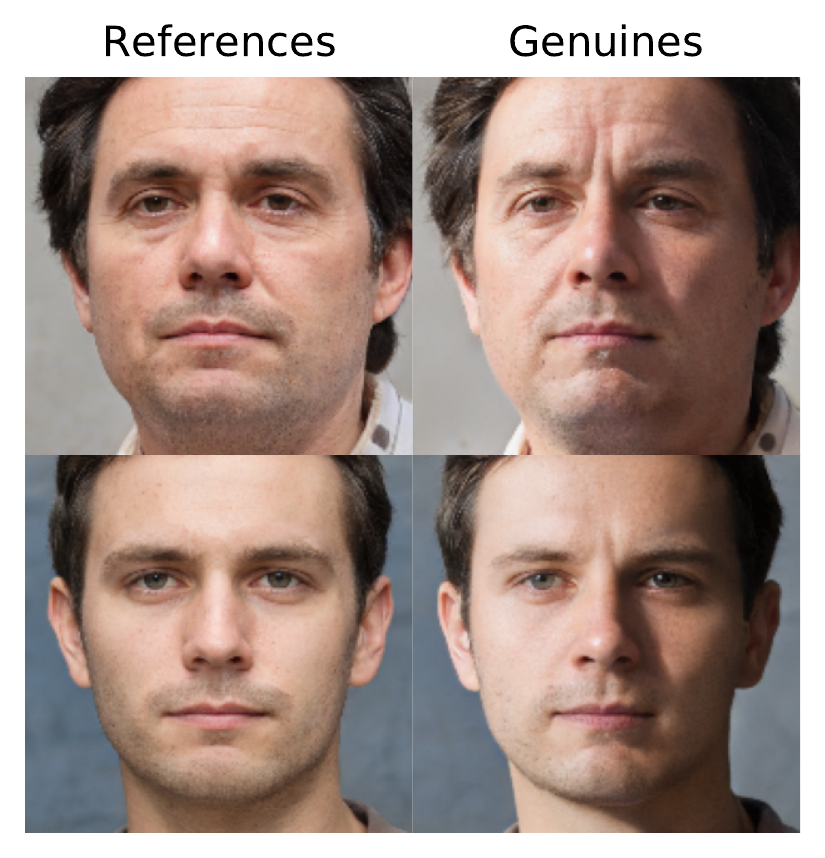}
		\caption{U - False non-matches}
		\label{fig:fnm_ir_U}
	\end{subfigure}
	\begin{subfigure}[b]{0.45\linewidth}
		\centering
		\includegraphics[width=\textwidth]{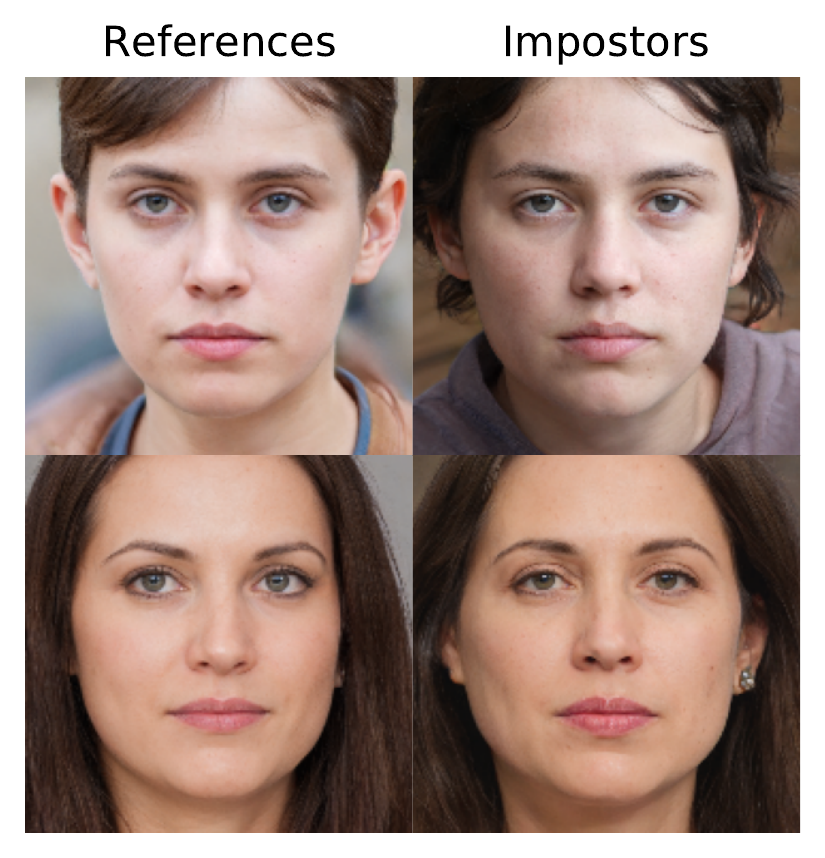}
		\caption{U - False matches}
		\label{fig:fm_ir_U}
	\end{subfigure}
	
	\begin{subfigure}[b]{0.45\linewidth}
		\centering
		\includegraphics[width=\textwidth]{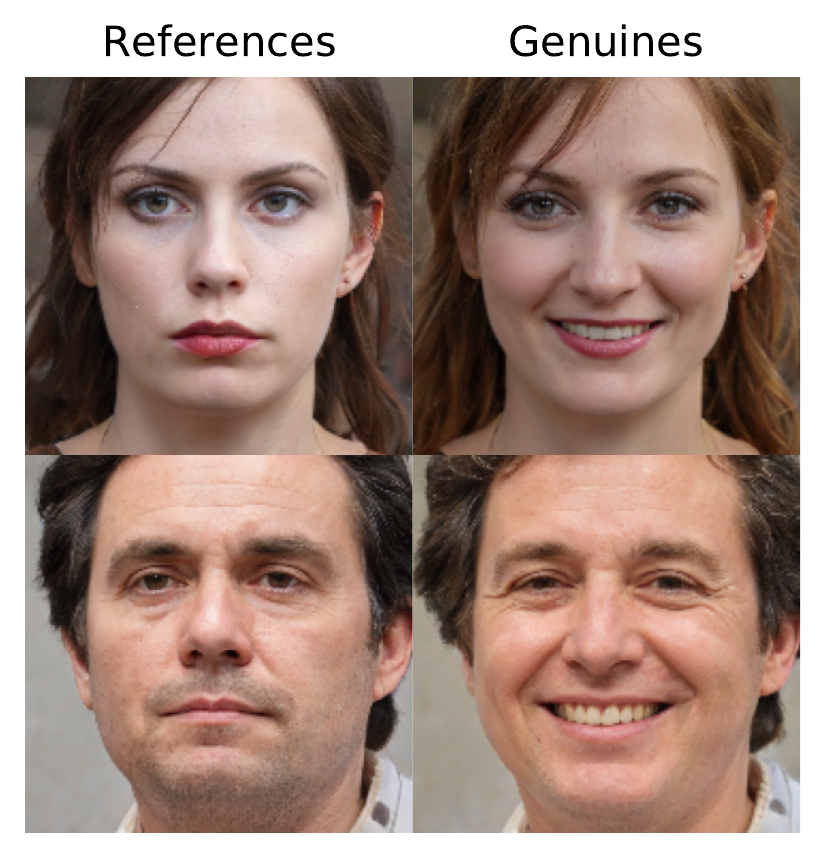}
		\caption{E - False non-matches}
		\label{fig:fnm_ir_E}
	\end{subfigure}
	\begin{subfigure}[b]{0.45\linewidth}
		\centering
		\includegraphics[width=\textwidth]{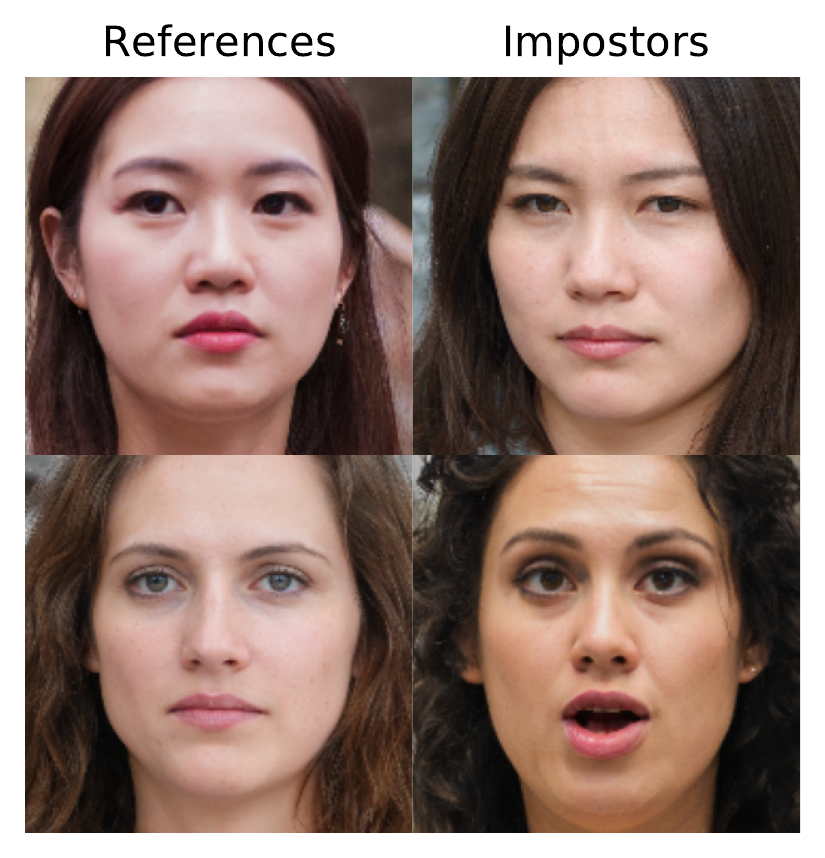}
		\caption{E - False matches}
		\label{fig:fm_ir_E}
	\end{subfigure}
	
	\begin{subfigure}[b]{0.45\linewidth}
		\centering
		\includegraphics[width=\textwidth]{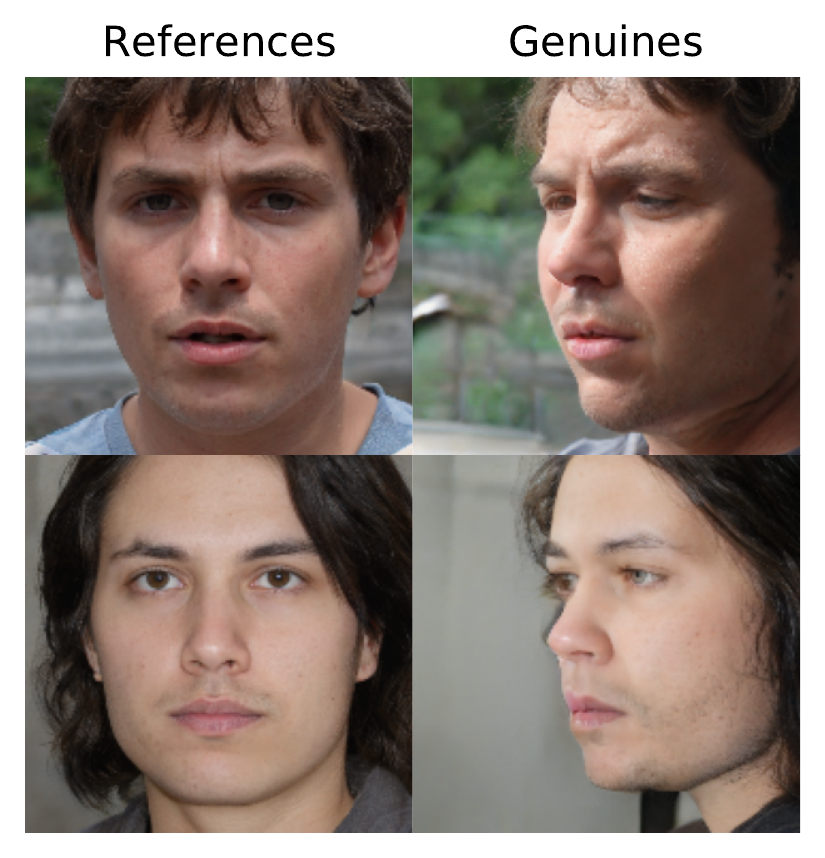}
		\caption{P - False non-matches}
		\label{fig:fnm_ir_P}
	\end{subfigure}
	\begin{subfigure}[b]{0.45\linewidth}
		\centering
		\includegraphics[width=\textwidth]{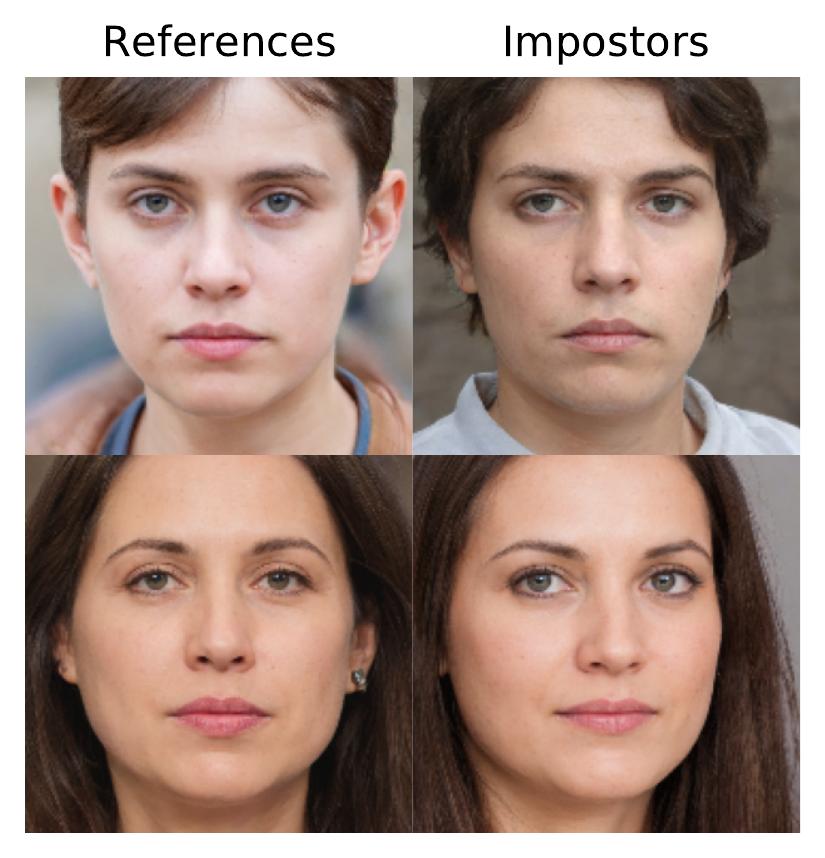}
		\caption{P - False matches}
		\label{fig:fm_ir_P}
	\end{subfigure}
	\caption{Worst matches at FMR@1E-3 with the Inception-ResNet v2 system evaluated on Syn-Multi-PIE. Additional examples are provided in the supplementary material.}
	\label{fig:matches_inception_resnet}
	\vspace{-0.5cm}
\end{figure}
\begin{table}
	\centering
	\begin{tabular}{|l|cc|cc|cc|}
		\hline
		{Protocol} & \multicolumn{2}{c|}{U} & \multicolumn{2}{c|}{E} & \multicolumn{2}{c|}{P} \\
		\hline
Compared & \footnotesize{MGS} & \footnotesize{SEP} & \footnotesize{MGS} & \footnotesize{SEP} & \footnotesize{MGS} & \footnotesize{SEP} \\
\hline\hline
Gabor                        &           0.17 &            1.6 &           0.05 &           1.11 &           0.09 &           1.19 \\
LGBPHS                              &           0.49 &           2.58 &           0.14 &           1.34 &           0.17 &           1.36 \\
AF           &           0.27 &            1.2 &          -0.22 &           0.97 &           0.21 &           1.27 \\
Inc-Res.v1&           0.52 &           0.85 &          -0.36 &            0.7 &           0.43 &           0.89 \\
Inc-Res.v2 &           0.32 &           1.06 &          -0.22 &           0.68 &           0.05 &           0.84 \\
AF-VGG2                      &           0.37 &           0.97 &           -0.4 &           0.68 &           0.16 &           0.92 \\
		\hline
	\end{tabular}
	\caption{Comparison of summary statistics of the score histograms. We report under MGS : $(\mu_{GS} - \mu_{GR})/|\mu_{GR}|$ (comparison of the Mean Genuine Similarity), and under SEP : $|\mu_{GS} - \mu_{IS}|/|\mu_{GR} - \mu_{IR}|$ (comparison of the mean SEParation between genuine and impostor scores), with $\mu_{GS}, \mu_{IS}$ resp. the mean Genuine \& Impostor scores for the Synthetic protocol, and $\mu_{GR}, \mu_{IR}$ the same but for the equivalent Real protocol. For MGS, any value $>0$ indicates a higher average similarity in genuine synthetic comparisons than in genuine real comparisons. For SEP, any value $>1$ indicates a stronger separation of the synthetic impostor \& genuine distributions than of the real ones.}
	\label{tab:hist_summary}
	\vspace{-0.5cm}
\end{table}
Table \ref{tab:fnmr} presents the measured error rates. In the ArcFace row for example (AF), the first 3 columns (Multi-PIE protocols) inform on the robustness of the system to each considered factor of variation. The system shows good robustness to illumination variation (U protocol, FNMR of 0.11), less robustness to expression variation (E protocol, FNMR of 0.3) and the least robustness to pose variation (P protocol, FNMR of 0.5). The last 3 columns provide the same information using Syn-Multi-PIE instead. We observe that our synthetic protocols are a good substitute to the real protocols, with a majority of Syn-Multi-PIE error rates lying at less than $5\%$ from the error rate on the corresponding Multi-PIE protocol. Qualitatively speaking, this means Syn-Multi-PIE satisfies the requirement of \emph{precision}, but there are still some imperfections. For illumination protocols the error rates follow a very similar trend, with the exception of the LGBPHS system. For pose protocols, there is a bit more discrepancy with the Inception-ResNet v2 and the ArcFace-VGG2 models, however the ranking of the systems remain very similar, with a clear performance improvement when using neural nets, and  with Inception-ResNet v2 appearing as the best model on this protocol. The picture is less clear on the expression protocols, with the Inception-ResNet and ArcFace-VGG2 models presenting significantly subpar performance on Syn-Multi-PIE. We now try to explain those failure cases. To do so, we study some visual examples of the worst false matches and non-matches in Syn-Multi-PIE protocols, in figure \ref{fig:matches_inception_resnet}. We especially focus on the Inception-ResNet v2 system, on which we also performed the identity uniqueness experiment in section \ref{sec:uniqueness}. Moreover, we also compare some summary statistics of the score histograms, which are presented in table \ref{tab:hist_summary}.

\subsection{Identity preservation}
We crucially want to check that we are correctly preserving identity during the latent editing process. Let us emphasize that the ``difficulty" of the synthetic protocols and hence the obtained error rates will be correlated with the strength of the editing. In particular, when we get a false non-match, is it because the FR system lacks robustness to the particular variation we are considering, or is it because the FR system is actually right and our image editing did not preserve the identity~? The highlighted discrepancies in table \ref{tab:fnmr} showcasing mainly \emph{excessive} error rates, it suggests we might indeed be in the situation of overly strong edits that do not preserve identity well. Those would indeed appear as false non matches in the analysis, as we still label those overly edited images as having the same identity as the reference. We thus consider the worst false non matches~: examples are presented in figures \ref{fig:fnm_ir_U}, \ref{fig:fnm_ir_E} and \ref{fig:fnm_ir_P}. 

Assessing preservation of identity is a subjective question. Readers can make their own opinion from the provided examples, but an extensive perceptual study would be necessary to answer this question more thoroughly. The examples at least suggest that identity preservation is good when doing illumination and expression editing. For pose editing however, the worst false matches contain enough perturbations to the facial details to possibly alter the identity. In this context our identity labeling could be partially erroneous, thus causing an increased number of false non matches. We could limit this issue by reducing the strength of the pose editing and renouncing to the most extreme angles. But the root cause might be an under-representation of extreme poses in the FFHQ dataset (and thus in StyleGAN2's output distribution) making the model fundamentally unable to generate far-from-frontal images.
To perform this analysis in a more objective manner, we can also read the Mean Genuine Similarity (MGS) comparison values in table \ref{tab:hist_summary}. An overly edited face image would generate an excessively low similarity score when compared to its reference. If our editing is too strong, we should thus observe a significant drop in the average synthetic genuine similarity score. In contrary, we observe that for the U and P protocol, the MGS is never lower in Syn-Multi-PIE than it is in Multi-PIE (all reported values are positive), which suggests that for both those covariates the editing is not too strong. The negative MGS comparison values with the E protocol might however hint towards slightly excessive expression editing.

\subsection{High similarity impostors}
Another possible cause of score discrepancy between synthetic and real protocols could be an excessive amount of very close synthetic identities causing a large number of high-score impostors. We showcase some of the worst false matches in figures \ref{fig:fm_ir_U}, \ref{fig:fm_ir_E} and \ref{fig:fm_ir_P}. We first notice that all those false matches are between feminine looking faces. This could be a possible indicator of a demographic differential of the Inception-ResNet v2 system against a feminine population, however it should then also occur during the Multi-PIE evaluation and it should not cause such a strong difference in performance. It can however be argued that some of the impostor identities actually do look very similar to the reference, to the point where a human judge might also consider them to be the same person. As seen in section \ref{sec:uniqueness}, the amount of variability between synthetic identities has been measured to be lower than in real data: we might be seeing the effect of this lack of variability. The summary statistics (SEP values in table \ref{tab:hist_summary}) lead to the same conclusion~: for neural-net based models, on E and P protocols, the distance between the impostor and genuine distributions is significantly smaller with Syn-Multi-PIE than it is with Multi-PIE. Even though we applied the ICT constraint when generating the references, it seems the subsequent face editions are enough to give way for many high score impostors, which are then responsible for the performance discrepancy w.r.t. the Multi-PIE benchmarks. In other terms, although for each synthetic identity we seem to span a realistic set of variations, the identities still are \emph{closer} from each other than real ones, and so there is more overlap between each of those identity-variation sets, causing an excessive amount of false matches. A natural way to fix this issue would be to increase the ICT value to spread out synthetic identities even more. But the issue that comes then is that with pure random sampling, we start to reject a high number of candidates when generating the reference identities, making the runtime grow quickly with the number of identities. Moreover, the ICT constraint is just a tentative to ``fix" the lack of variability of StyleGAN2 identities. We can hope future improvements in face GANs will further increase the richness of their output distribution, in particular identity variability.

\section{Conclusion}
In this work, we have presented a method to generate synthetic databases for face recognition by exploiting the StyleGAN2 latent space. This has enabled us to introduce the Syn-Multi-PIE dataset,  which presents the same factors
of variation as the Multi-PIE database, but does not contain any real identity and thus can be shared much less restrictively. Moreover, our database construction process is automatic and can be extended to other covariates.
Benchmark experiments
carried out with 6 face recognition systems have shown that Syn-Multi-PIE can be used in place of Multi-PIE and lead to similar conclusion on the systems' error rates and ranking.  
While we have noticed some performance discrepancies in some of the setups, our analysis suggests those discrepancies generally do not seem to be caused by our editing method, which looks to preserve identity quite well. We do note that our quantitative observations (MGS) on expression editing raise some doubt on the quality of the identity preservation in this context, however the visual examples (false non-matches) reinforce the claim. In a future study, it could be useful to perform a perceptual study with human raters, as it remains the only way to fully validate identity preservation. 

Overall, performance discrepancies seem rather caused by limitations of the generative capacity of StyleGAN2, mainly the lack of variability of generated identities, and a lack of extreme poses in the bulk of the output distribution. The field of generative face models currently following a very fast evolution, we expect those issues to improve as new models emerge able to produce an even more realistic output image distribution. 

Other usages for synthetic data can be conceived. Given the available semantic control in the $\W$ space, one could for example generate datasets with a balanced density of each demographics (gender, age or race), which could be of use to expose possible demographic differentials in current SOTA systems. Finally, our generation process scaling at least to 10k+ identities, a natural next step would be to generate a large-scale dataset and assess its usability for \emph{training} FR models while solving the data privacy issue. This direction has indeed not been deeply explored, at least using GAN-based generative models. Again, it would probably require a stronger preliminary verification of the identity preservation at the editing step, else we can foresee mislabeled synthetic images causing issues during the training, leading to subpar performance.  

\vspace{-0.25cm}
{\small
\bibliographystyle{ieee}
\bibliography{mybib}
}

\clearpage
\pagenumbering{arabic}
\appendix
\section{Database generation details}
In this section we provide a bit more technical details on the synthetic database generation process.
First, we describe the database projection protocol, which for each covariate of interest specifies which subset of images we project to form the two latent populations on which the SVM is fitted. We then provide some details on the generation of reference images, in particular on the face neutralization process, including a pseudocode algorithm. Finally, we present in a second pseudocode the process of generating variations for each identity, using the precomputed latent directions. 

\subsection{Database projection protocol}
We project the Multi-PIE database \cite{multipie}. For each identity, it contains face images with labeled covariates for expression, view angle and lightning direction. The available expressions are \emph{neutral, smile, disgust, scream, squint, surprise}. For notation convenience, we number them in that order from 0 to 5.

We project the world set of Multi-PIE, then use the projected latents to fit SVMs in $W$ and find interesting latent directions. Table \ref{tab:multipie_subsets} specifies our protocol, i.e. which subset of the projected latents are used to compute each of the latent directions of interest. After this process we have access to the following latent directions (unit vector normals to the SVMs hyperplanes) : left-right pose edition ($\dir{P}$), left-right illumination edition ($\dir{I}$), and edition between any pair of expressions, for example $\dir{01}$ for neutral-to-smile and $\dir{02}$ for neutral-to-disgust.

We also need some notion of scaling, i.e. to determine how much we can move along a given latent direction while still preserving the identity and realism of the face. As a reasonable heuristic, we thus also keep track of the mean distance to the hyperplane of each population of projected latents. We denote it by $d_P^L$, $d_P^R$ for the left and right pose, $d_I^L$,$d_I^R$ for the left and right illumination, and by $d_{01}^0$ and $d_{01}^1$ for example for the neutral and smile populations with respect to the $\dir{01}$ hyperplane.

\begin{table*}
	\begin{center}
		\begin{tabular}{|l|c|c|c|c|}
			\hline
			Attribute & Expressions & Cameras & Lightning & Binary classes \\
			\hline\hline
			Expression & All & Frontal view & Frontal flash & All expression pairs \\
			Pose & Neutral & Non frontal views in $[-45\degree, 45\degree]$ & Frontal flash & Left \& right profiles \\
			Illumination & Neutral & Frontal view & All non frontal with flash & Left \& right flashes\\
			\hline
		\end{tabular}
	\end{center}
	\caption{Description of the Multi-Pie subsets used to compute the latent directions for each attribute.}
	\label{tab:multipie_subsets}
\end{table*}

\subsection{Generation of identities}
\begin{figure}[t]
	\centering
	\includegraphics[width=0.95\linewidth]{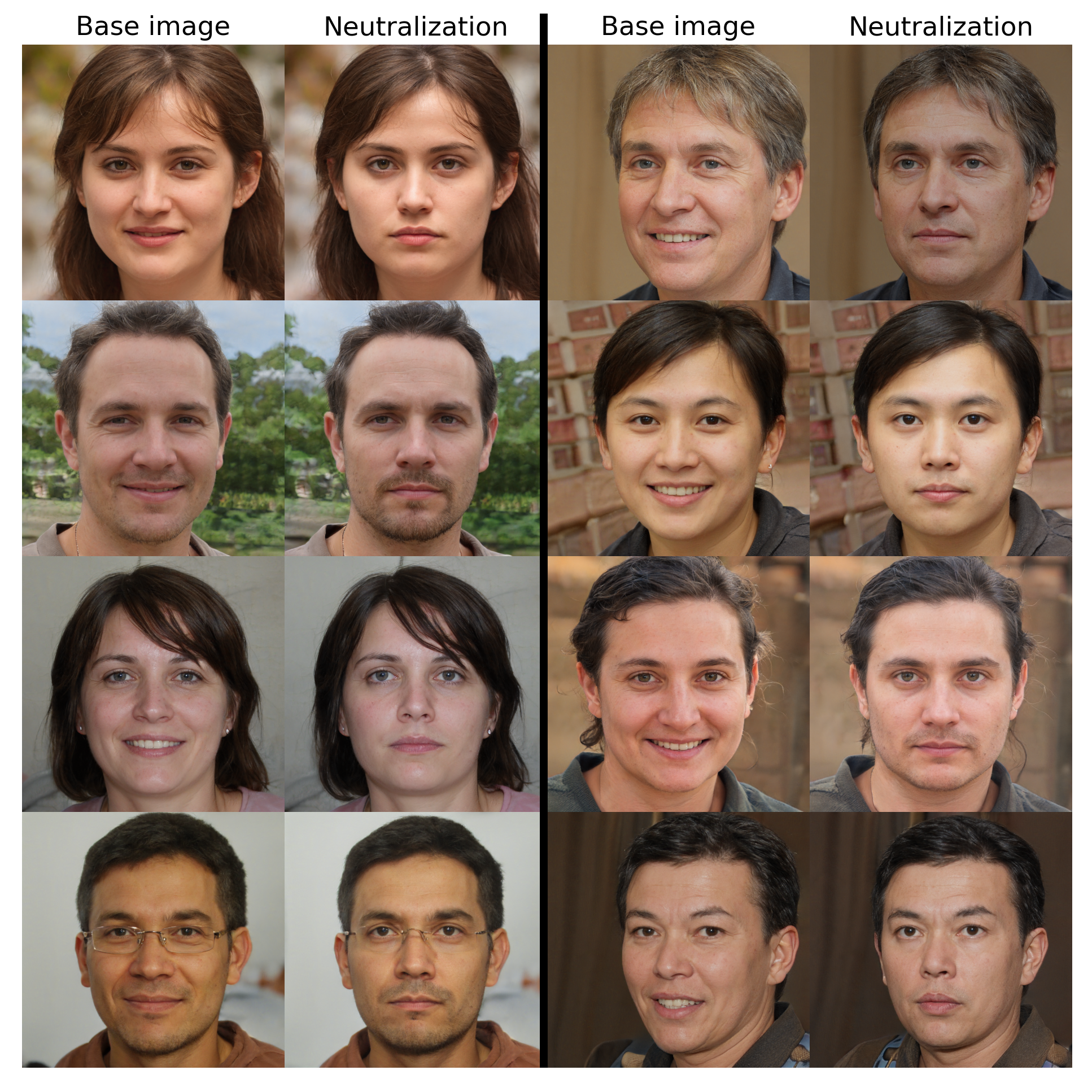}
	\caption{Examples of generated faces before and after the face neutralization process. While the expression neutralization seems qualitatively successful, it is not the case for pose neutralization. But we do not need the faces to be perfectly frontal : the only aim is to make sure to keep reasonable poses when editing the image along the pose editing direction, so we simply need to start close to a frontal view. Illumination neutralization is not very visible, due to the absence of extreme illuminations in the typical uncontrolled output of StyleGAN2.}
	\label{fig:faceneutralization}
\end{figure}
Algorithm \ref{alg:references} explains in pseudocode the process of generating the references, using our notation from before for the latent directions. It includes a neutralization step, during which the candidate face is edited to have frontal view, frontal lightning and neutral expression. After neutralization, candidate faces are optionally rejected based on the comparison of their embedding distance to all previous faces w.r.t the required ICT. Figure \ref{fig:faceneutralization} presents example of generated faces before and after the neutralization process.

\begin{algorithm}
	\footnotesize
	\caption{Creation of reference identities}
	\begin{algorithmic}
		\Procedure{New\_ID}{$W_{prev}$, ICT}  	
		\Repeat
		\State $\mathbf{z} \leftarrow \text{Random z-latent}$
		\State $\mathbf{w} \leftarrow \text{\textsc{Mapping}}(\mathbf{z})$ \Comment{StyleGAN2 mapping}
		\State $\mathbf{w_{ref}}$ $\leftarrow{\text{\textsc{Neutralize}}(\mathbf{w})}$
		\Until{\textsc{ClosestDistance}($\mathbf{w_{ref}}$, $W_{prev}$) $>$ ICT}
		\State \textsc{Append}($W_{prev}$, ref)
		
		\EndProcedure
		
		\State
		\Function{Neutralize}{$\mathbf{w}$}
		\State $\mathbf{w} \leftarrow \mathbf{w} - (\mathbf{w}
		^\top \dir{P}) \cdot \dir{P} $ \Comment{Pose}
		\State $\mathbf{w} \leftarrow \mathbf{w} - (\mathbf{w}
		^\top \dir{I}) \cdot \dir{I} $ \Comment{Illumination}
		
		\State $\mathbf{w} \leftarrow \mathbf{w} - (\mathbf{w}
		^\top \dir{01} + d_{01}^0) \cdot \dir{01}$ \Comment{Expression}
		\State \Return $\mathbf{w}$
		\EndFunction
		\State
		\Function{ClosestDistance}{$\mathbf{w}$, $W_{prev}$}
		\State $\mathbf{e} = \text{\textsc{Embedding}(\textsc{Synthesis}}(\mathbf{w}))$ 
		\Comment{Compute image and face embedding}
		\State dists $\leftarrow$ []
		\ForAll{$\mathbf{w'}$ in $W_{prev}$}
		\State $\mathbf{e'} = \text{\textsc{Embedding}(\textsc{Synthesis}}(\mathbf{w'}))$
		\State \textsc{Append}(dists, $\text{\textsc{CosineDistance}}(e, e')$)
		\EndFor
		\State \Return \textsc{Minimum}(dists)
		
		\EndFunction
	\end{algorithmic}
	\label{alg:references}
\end{algorithm}

\subsection{Generation of variations}
Algorithm \ref{alg:variations} presents in pseudocode the process of generating variations for each reference, using the precomputed latent editing directions.

\begin{algorithm}
	\footnotesize
	\caption{Creation of variations}
	\begin{algorithmic}
		\Procedure{Variations}{$\mathbf{w_ref}$, $n_{var}$}
		\State pose\_var $\leftarrow$ \textsc{LRVar}($\mathbf{w_{ref}}$, $\dir{P}$, $d_P^L$, $d_P^R$, $n_{var}$)
		\State illum\_var $\leftarrow$ \textsc{LRVar}($\mathbf{w_{ref}}$, $\dir{I}$, $d_I^L$, $d_I^R$, $n_{var}$)
		\State expr\_var $\leftarrow$ \textsc{ExprVar}($\mathbf{w_{ref}}$)
		
		\EndProcedure	
		\State
		\State 
		\textbf{Left-Right editing}
		\Function{LRVar}{$\mathbf{w}$, $\dir{}$, $d^L$,
			$d^R$, $n_{var}$} 
		\State variations $\leftarrow$ []
		\State $D\leftarrow$ \textsc{Maximum}($d^L$, $d^R$)
		\State dists $\leftarrow$ \textsc{LinSpace}($-D$, $D$,$n_{var}$)
		\ForAll{$d$ in dists}
		\State \textsc{Append}(variations, $\mathbf{w} + d\cdot \dir{}$)
		\EndFor
		\State \Return variations
		\EndFunction
		
		\State
		\State \textbf{Expression editing}
		\Function{ExprVar}{$\mathbf{w}$}
		\State variations $\leftarrow$ []
		\For{$j$ in 1..5}
		\State \textsc{Append}(variations, $\mathbf{w} + (-\mathbf{w}^\top \dir{0j} + d_{0j}^j)\cdot\dir{0j}$)
		\EndFor
		\EndFunction
	\end{algorithmic}
	\label{alg:variations}
\end{algorithm}

\section{Visual examples}
\begin{figure}
	\centering
	\begin{subfigure}[b]{0.45\linewidth}
		\centering
		\includegraphics[width=\textwidth]{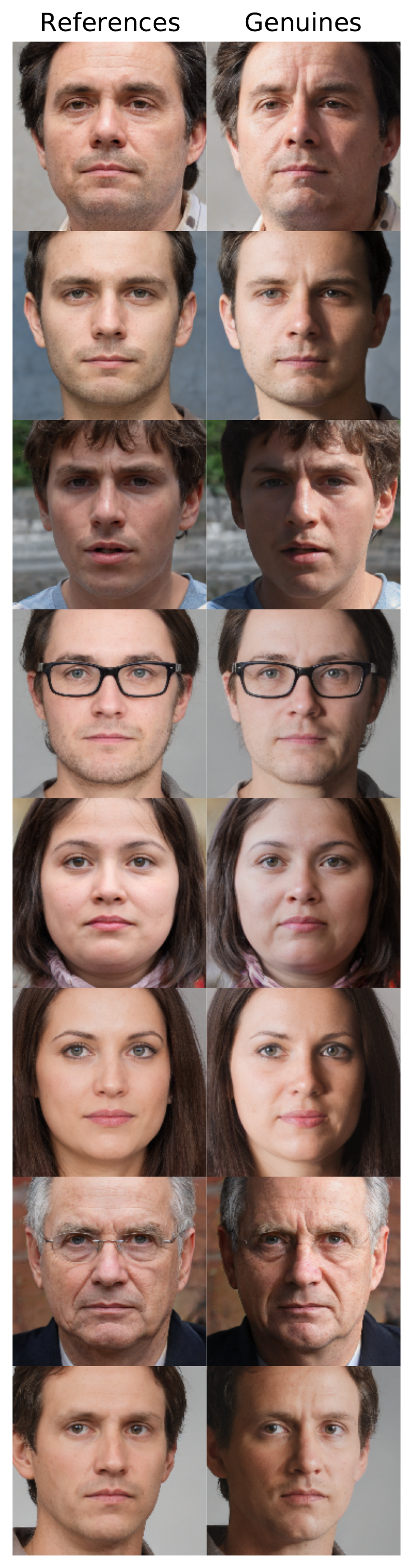}
		\caption{U - False non-matches}
		\label{fig:fnm_ir_U_plus}
	\end{subfigure}
	\begin{subfigure}[b]{0.45\linewidth}
		\centering
		\includegraphics[width=\textwidth]{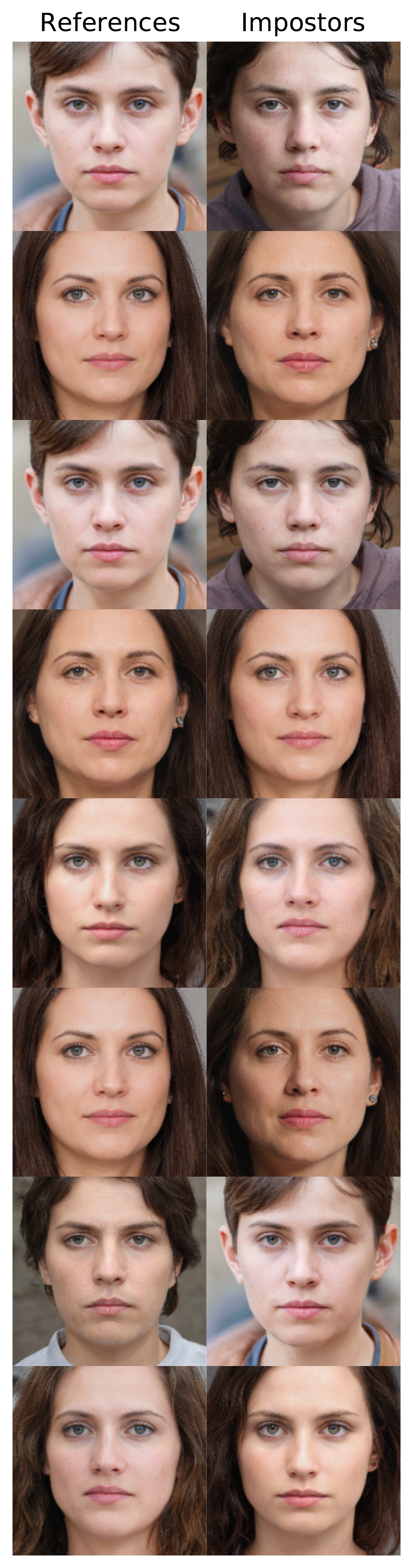}
		\caption{U - False matches}
		\label{fig:fm_ir_U_plus}
	\end{subfigure}
	\caption{Worst matches at FMR@1E-3 with the Inception-ResNet v2 system evaluated on the Syn-Multi-PIE U protocol.}
	\label{fig:matches_inception_resnet_U_extended}
\end{figure}

\begin{figure}
	\centering
	\begin{subfigure}[b]{0.45\linewidth}
		\centering
		\includegraphics[width=\textwidth]{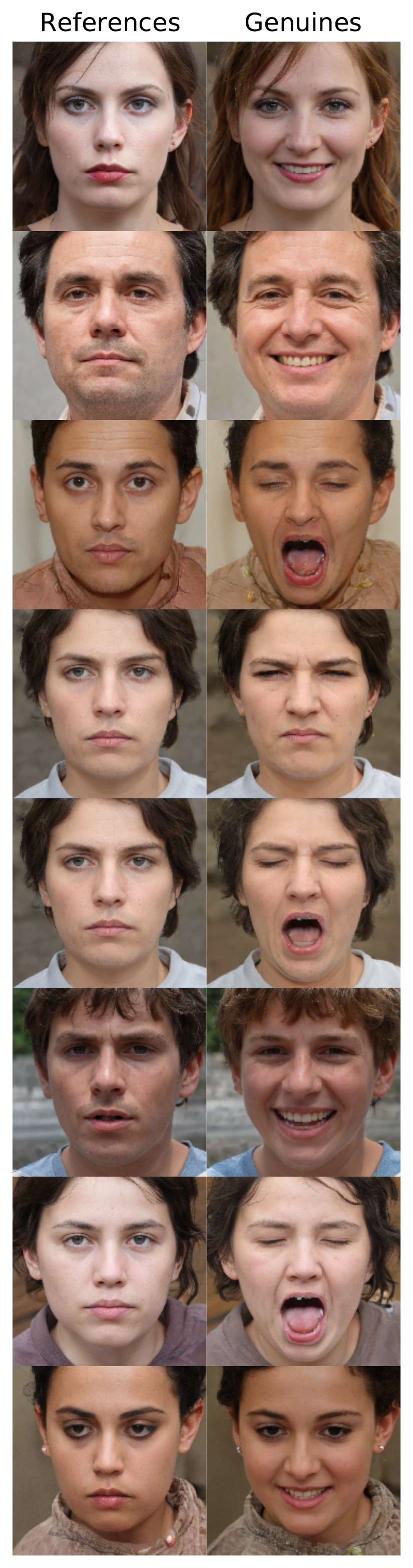}
		\caption{E - False non-matches}
		\label{fig:fnm_ir_E_plus}
	\end{subfigure}
	\begin{subfigure}[b]{0.45\linewidth}
		\centering
		\includegraphics[width=\textwidth]{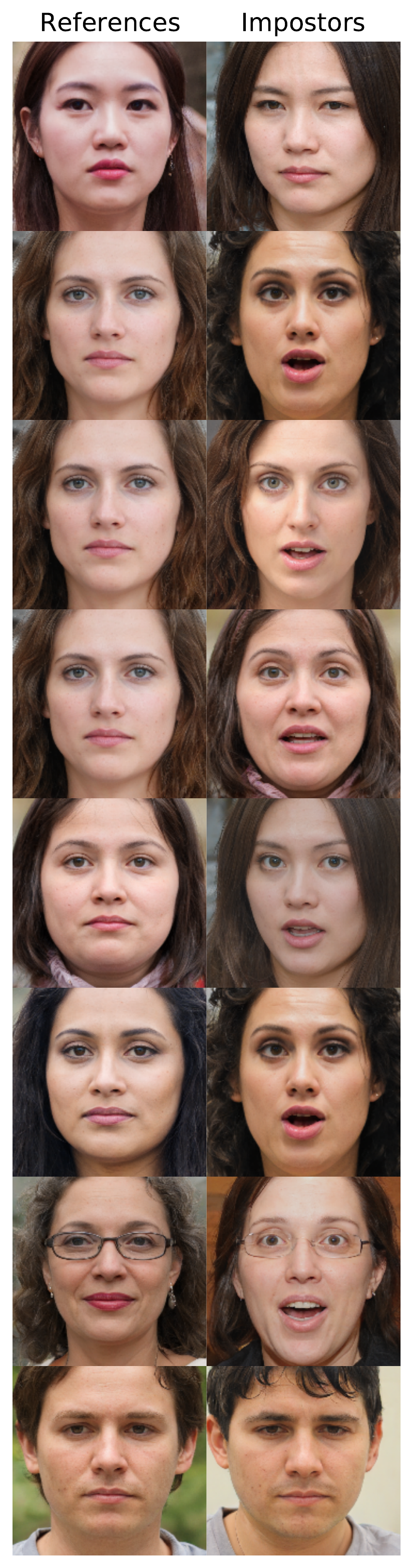}
		\caption{E - False matches}
		\label{fig:fm_ir_E_plus}
	\end{subfigure}
	\caption{Worst matches at FMR@1E-3 with the Inception-ResNet v2 system evaluated on the Syn-Multi-PIE E protocol.}
	\label{fig:matches_inception_resnet_E_extended}
\end{figure}

\begin{figure}
	\centering	
	\begin{subfigure}[b]{0.45\linewidth}
		\centering
		\includegraphics[width=\textwidth]{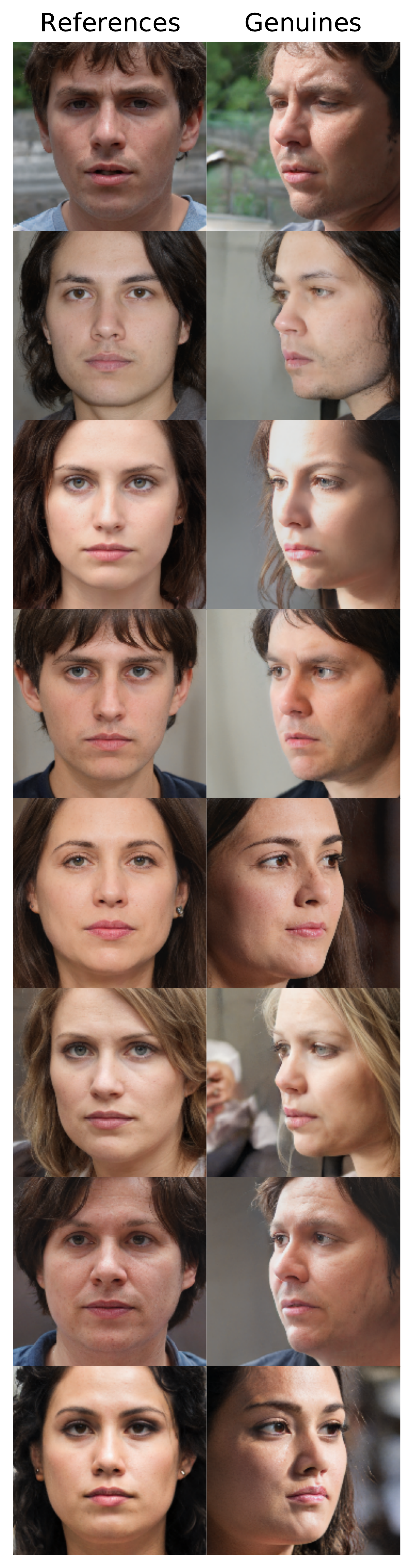}
		\caption{P - False non-matches}
		\label{fig:fnm_ir_P_plus}
	\end{subfigure}
	\begin{subfigure}[b]{0.45\linewidth}
		\centering
		\includegraphics[width=\textwidth]{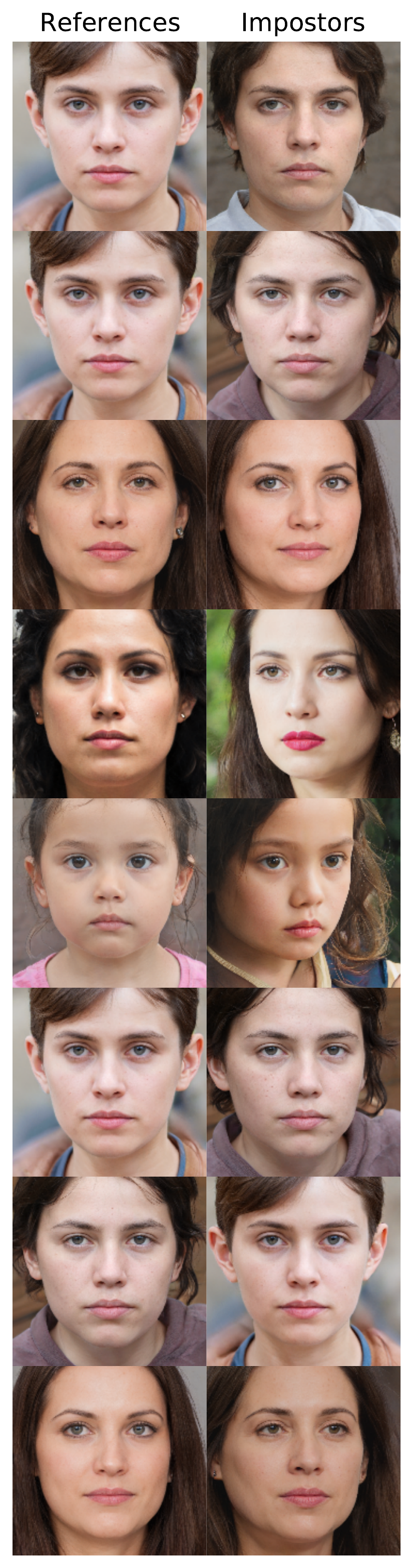}
		\caption{P - False matches}
		\label{fig:fm_ir_P_plus}
	\end{subfigure}
	\caption{Worst matches at FMR@1E-3 with the Inception-ResNet v2 system evaluated on the Syn-Multi-PIE P protocol.}
	\label{fig:matches_inception_resnet_P_extended}
\end{figure}

In this section, we provide some more visual examples of false matches / non-matches obtained at FAR@1E-3 when evaluating the Inception-ResNet v2 system on the Syn-Multi-PIE protocols, respectively U, E and P in figures \ref{fig:matches_inception_resnet_U_extended}, \ref{fig:matches_inception_resnet_E_extended} and \ref{fig:matches_inception_resnet_P_extended}.

\end{document}